# A Hybrid Architecture with Efficient Fine Tuning for Abstractive Patent Document Summarization

Nevidu Jayatilleke
School of Computing
Informatics Institute of Technology
Colombo, Sri Lanka
nevidu.20200878@iit.ac.lk

Ruvan Weerasinghe
School of Computing
Informatics Institute of Technology
Colombo, Sri Lanka
ruvan.w@iit.ac.lk

***Abstract*—Automatic patent summarization approaches that help in the patent analysis and comprehension procedure are in high demand due to the colossal growth of innovations. The development of natural language processing (NLP), text mining, and deep learning has notably amplified the efficacy of text summarization models for abundant types of documents. Summarizing patent text remains a pertinent challenge due to the labyrinthine writing style of these documents, which includes technical and legal intricacies. Additionally, these patent document contents are considerably lengthier than archetypal documents, which complicates the process of extracting pertinent information for summarization. Embodying extractive and abstractive text summarization methodologies into a hybrid framework, this study proposes a system for efficiently creating abstractive summaries of patent records. The procedure involves leveraging the LexRank graph-based algorithm to retrieve the important sentences from input parent texts, then utilizing a Bidirectional Auto-Regressive Transformer (BART) model that has been fine-tuned using Low-Ranking Adaptation (LoRA) for producing text summaries. This is accompanied by methodical testing and evaluation strategies. Furthermore, the author employed certain meta-learning techniques to achieve Domain Generalization (DG) of the abstractive component across multiple patent fields.**

## I. INTRODUCTION

Text summarization is a widely recognized task within the field of NLP. The emphasis on text summarization has increased significantly due to the rapid growth of various text-generating activities. Condensing lengthy papers has positively influenced several aspects across many fields and domains. There are several advantages to reading summarized material as opposed to lengthy descriptions. These include a better grasp of the most pertinent information, a reduction in the time required for reading, and a decrease in cognitive load due to the presentation of information in a more easily comprehensible style.

The endeavor of text summarization can be categorized into two distinct approaches: extractive and abstractive. Extractive summarization employs a technique that selects a subset of words, phrases, or sentences from the original content to create a summary. Abstractive summarization, on the other hand, generates sentences that succinctly overview the material while capturing the essential concepts and aspects of the source text. This approach typically involves making significant modifications and paraphrasing the original sentences to achieve the desired summary. Generally, extractive summarization entails highlighting the most important information from the entire text. In contrast, abstractive summarization focuses on capturing key information and producing a summary that more closely resembles human language. This study primarily centers on abstractive summarization, emphasizing the incorporation of certain extractive strategies to enhance performance as the main focus of the author's investigation.

Patents are a type of intellectual property that gives people or organizations ownership and exclusive rights. They are publicly available documents, often containing illustrations, that are registered by governments and international organizations. The registration process makes it possible for those knowledgeable about the domain to understand how to reproduce important and inventive innovations. Nevertheless, it restricts production unless the patent owner issues a license or enters into a formal agreement to transfer ownership [1]. The important information in patent documents is dispersed evenly throughout the document, which makes summarizing much more difficult even when human cognition is present [2]. The growing number of inventions over time has led to a high demand for efficient document organization while maintaining the importance of these documents. Performing a summarization task would assist in achieving this goal.

Comprehensive research on resource-efficient methodologies remains inadequate despite notable progress in using AI systems to create abstractive text summaries for patent documents. It typically consumes extensive computational power to train and fine-tune models for processing long documents. Existing research shows that complex systems and architectures trained over several days on high-end computer resources yield reasonable outcomes in terms of assessment scores, but computationally cost-effective methods yield noticeably less effective outcomes. Consequently, models that are economical in their use of resources and capable of summarizing patent documents are required to bridge the gap between current capabilities and the anticipated level of competence.

In this study, the author presents the findings and implementations of the research, spanning from the start of the literature review to the final architectural design. The next section provides an overview of a comprehensive investigation into the currently available systems for abstractive summarization of patent documents. Furthermore, the proposed hybrid architecture, designed to address the identified gap, is illustrated and explained thoroughly in the following segments. Additionally, the generalization of the model, conducted using few-shot learning to generate summaries from unseen data across selected domains is discussed to demonstrate its effectiveness. Most importantly, the evaluation and testing methodologies used to determine whether the author has achieved the expected results are presented with clarity. Finally, the author shares potential future work aimed at enhancing this study to improve patent and legal document summarization significantly. The findings and methodologies presented in this section serve as a cornerstone for advancing abstractive summarization in specialized domains.

## II. Existing Work

Patent document summarizing is a prevalent field of study that has been explored utilizing numerous techniques and methodologies. This study reviews several abstractive text summarization approaches that have demonstrated remarkable progress.

A framework for summarizing is proposed, including a learning stage based on a GAN architecture. The GAN-based summarization model features a generator and a discriminator that compete against each other. The generator is a transformer-based model designed for text summarization. It generates summary sentences by taking individual patent text data and labeling data as input. The generator comprises an encoder and a decoder. The construction of the transformer involves setting up stacked encoder and decoder blocks. In this study, a total of four blocks were stacked and utilized. The encoder block of the generator includes an embedding layer, a multi-head attention layer, a first regularization layer, a feed-forward layer, and a second regularization layer. After inputting the complete text into the encoder, the layer operations are executed sequentially, calculating the encoder block output. At this stage, the transformer model applies positional encoding to combine word embedding and location information of each word. The decoder consists of a masked multi-head attention layer, a first regularization layer, a multi-head attention layer, a second regularization layer, a feed-forward layer, and a final regularization layer. The word produced in the previous stage is fed into the decoder's first layer, while the encoder block's output is input into the multi-head attention layer of the decoder block. The output of the decoder block is ultimately sent to the dense layer, which then generates the next word.

The discriminator is a model that takes in the textual data of summary sentences and assesses whether each sentence is a created or an actual sentence. In this work, the discriminator is a classification model with a single Bi-LSTM layer. It receives the generated summary sentence or the target summary sentence as input and verifies its authenticity, distinguishing between actual and generated sentences. The discriminator's output serves as a reward to train the generator and discriminator [3].

A different approach comprises two distinct trainable components. An extractive model, comprising a hierarchical encoder that outputs sentence representations, is used to either point to or classify sentences in the input and a transformer language model, conditioned on the extracted sentences as well as a part of or the entire input document. A single transformer language model with 220M parameters, 20 layers, 768-dimensional embeddings, 3072-dimensional position-wise Multi-Layer Perceptrons (MLPs), and 12 attention heads has been constructed and trained from scratch. To enable an unconditional language model to perform abstractive summarization, they used the fact that language models are trained by decomposing the joint distribution of words in an autoregressive manner. Simply put, they usually factorize the joint distribution of tokens into a product of conditional probabilities. Consequently, they structure the training data for the models to ensure the ground-truth summary follows the information utilized by the model to produce a summary [4] [5].

A distinct method presented a new model called LongT5, which was applied to study the joint impact of changing the input length and model size. They do this by incorporating pre-training and long-input transformer attention into the scalable T5 model [6] design. The fundamental concept is to encourage the model to produce significant phrases from a document as a single string, much like a summary, by selectively masking those sentences. From an architectural standpoint, the attention mechanism is the primary difference between T5 and LongT5. For LongT5, they investigate two different kinds of attention mechanisms: transient global attention (TGlobal) and local attention. Both versions preserve several T5 features, such as compatibility with T5 checkpoints, support for example packing, and relative position representations. Its ability to handle long inputs, which makes it ideal for summarizing tasks, was the main factor in this pre-trained model's improved performance which was fine-tuned using many datasets [7].

Transformers, like BERT, have become very successful deep-learning models for NLP. However, one of the main problems with these models is that they have a quadratic dependency, especially in memory, on the sequence length due to their full attention mechanism. To solve this problem, a study suggested BIGBIRD, which has a sparse attention mechanism that effectively reduces the quadratic dependency to linear. It is shown that BIGBIRD preserves these features of the quadratic, full-attention model by acting as a universal approximator of sequence functions and being Turing complete. Their theoretical research reveals the benefits of global tokens, which consider the entire sequence as part of the sparse attention mechanism. Sequences up to eight times longer than previously possible with comparable strategies can be processed using the proposed sparse attention mechanism [8].

In a recent study, a novel Sequence to Sequence with Attention (SSWA) model was presented, which comprises an encoder and a decoder. The input sequences of a patent document, including the title, abstract, and claim, are converted into a preset length using a zero-padding technique before being fed into the encoder, which uses vector data to represent the patents. The decoder uses the target sequence (summary) and the previous encoding representation to predict the next summary words. The attention mechanism focuses attention on sequences supplied by the encoder for each decoded word in the decoder. The research employed a bi-directional long and short-term memory model as the encoder to extract hidden features from a sentence's word embedding and an LSTM model to predict the output word in the decoder step. The attention mechanism in the decoder records the hidden states before word generation to give each decoded word a higher priority [1].

The importance of patent document summarization is evident from the large amount of research that has been done on its successful application, and it is also logical to conclude that transformer-based architectures are being used widely in this field. Nevertheless, the current work has not addressed the resource efficiency problem that was previously discussed with structured approaches, which emphasizes the identified research gap.

## III. Proposed Hybrid Architecture

The suggested system consists of three primary components: data preprocessing, sentence ranking, and abstractive text summarization. Furthermore, an overview of training data augmentation for few-shot learning is included in this section. Together, these components form a cohesive framework designed to enhance the performance and adaptability of abstractive summarization systems.

### A. Sentence Ranking Algorithm

The sentence ranking component utilized the LexRank algorithm, which leverages graph theory to rank sentences based on their importance. It was then fed to the text summarization model to generate the final summary.

Graph-based methods utilize graph theories to create text summarization techniques. Following standard preprocessing techniques like stemming and stop word removal, sentences within the documents are depicted as nodes in an undirected graph. Sentences are linked by edges based on sentence structure [9]. Such representation is frequently utilized for extractive summarization in LexRank and TextRank [10].

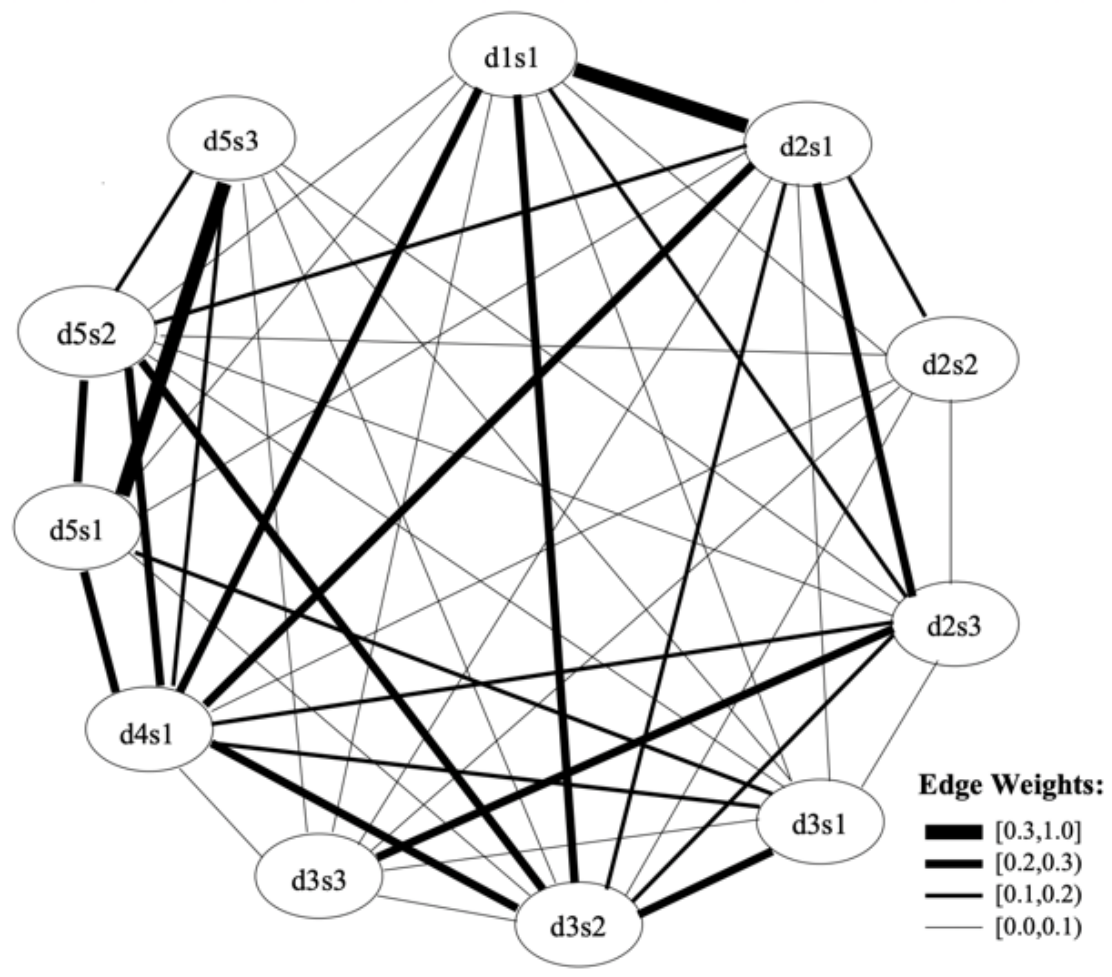

Fig. 1. Weighted cosine similarity graph representing inter-sentence relationships across multiple documents. Nodes represent individual sentences, and edge weights reflect the cosine similarity between sentence embeddings. (e.g., d1-s1 denotes sentence 1 from document 1) [11].

In the LexRank algorithm, a sentence that is similar to multiple other sentences in the text would probably be considered important. The method recommends a specific sentence based on its similarity to other sentences, resulting in a higher ranking. This method is based on the Eigenvector Centrality. It adheres to a connected graph approach. Each sentence is positioned at a vertex of the graph. The weight on the edges is determined using a cosine similarity metric, as shown in Fig 1 [12].

In this study, the PageRank algorithm is imported directly through the NetworkX library, and the necessary modifications were made to develop the LexRank algorithm accordingly. The similarity graph for sentences in LexRank is undirected because cosine similarity, used to measure similarity, is a symmetric relation. This contrasts with the original PageRank approach, which was designed to compute web page prestige. The formula provided below offers a clear overview of the LexRank algorithm that utilizes weighted graphs to rank the sentences.

$$p(u) = \frac{d}{N} + (1-d)\sum_{v \in adj[u]} \frac{idf\!-\!modified\!-\!cosine(u,v)}{\sum_{z \in adj[u]} idf\!-\!modified\!-\!cosine(z,v)} p(v) \quad (1)$$

where p(u) is the centrality of node u, adj[u] is the set of nodes that are adjacent to u, N is the total number of nodes in the graph, and d is a “damping factor” [13].

The author employed count vectorization to create word vectors for this component in the study. Based on the experimentation carried out, it was determined that a damping factor of 0.85 and a similarity threshold for voting of 0.2 should be selected.

The author identified that the LexRank algorithm is derived from the TextRank algorithm and can be considered an improved version of it. Significant differences exist between the two algorithms, which are then followed by common traits. However, there are significant differences in the sentence ranking between LexRank and TextRank, especially in the use of cosine similarity to measure word overlaps across sentences.

### B. Abstractive Text Summarization Model

After the sentence ranking is completed, the system moves on to the second and most crucial phase: the abstractive text summarization model. During this phase, the system tokenizes the ranked sentences and inputs the first 1024 tokens into the model, aiming for an output with a maximum sequence length of 512 tokens.

The author has decided to use pre-trained transformers and modify them, particularly for the summarizing task. An extensive literature review revealed that Parameter Efficient Fine Tuning (PEFT) is an extremely successful and effective transformer fine-tuning methodology. Among the various PEFT approaches that are accessible, Low-Rank Adaptation (LoRA) is the fine-tuning technique that has been selected for the suggested design. Its efficacy and recent extensive recognition significantly influenced this strategy's selection. By optimizing the rank-decomposition matrices of the dense layers' change during adaptation, the LoRA approach essentially makes it possible to train dense layers in a neural network. As shown in Fig. 2, this is carried out with the pre-trained weights frozen.

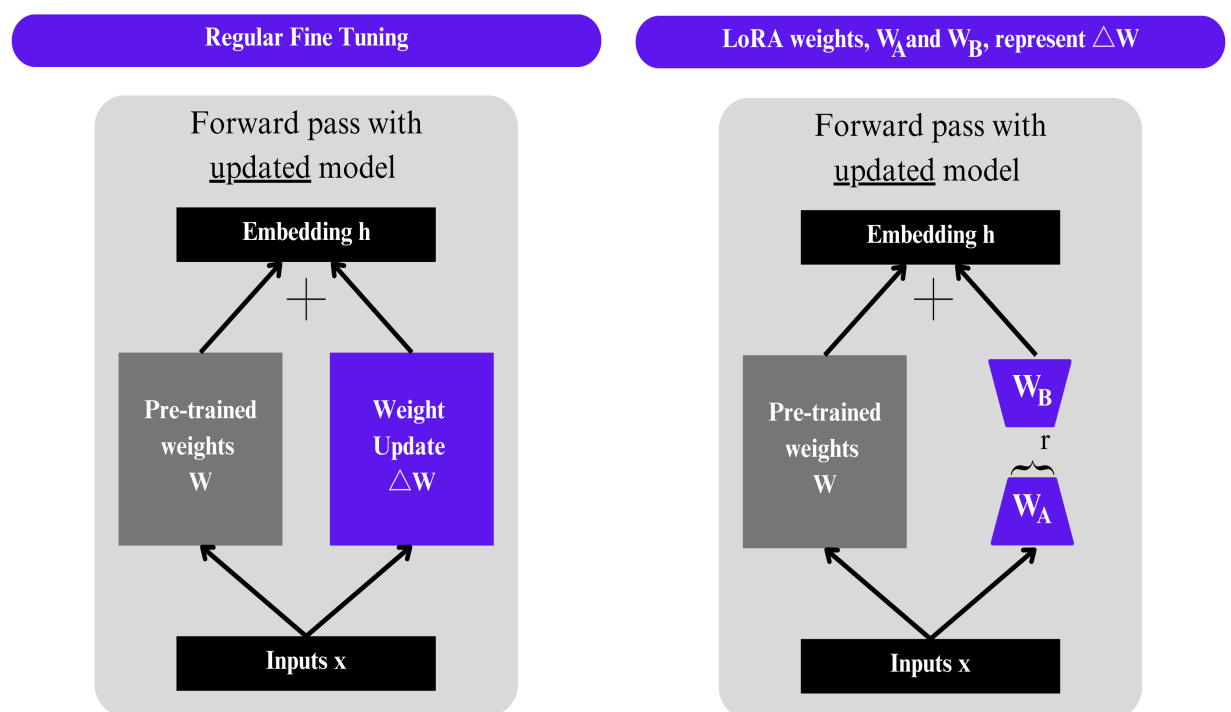

Fig. 2. Comparison of parameter updates during LoRA reparameterization and regular fine-tuning.

To constrain the update of a pre-trained weight matrix $W_0 \in R^{d \times k}$, its update is represented by a low-rank decomposition $W_0 + \Delta W = W_0 + BA$, where $B \in R^{d \times r}$, $A \in R^{r \times k}$, and the rank $r \ll \min(d,k)$. Although A and B contain trainable parameters, $W_0$ is fixed and does not receive gradient updates during training. Both $W_0$ and $\Delta W = BA$ are multiplied by the same input, and their respective output vectors are aggregated coordinate-wise. When $h = W_0 x$, the modified forward pass gives us the following result:

$$h = W_0 x + \Delta W x = W_0 x + BAx \quad (2)$$

At the start of training, utilize a random Gaussian initialization for matrix A and set matrix B to zero. This ensures that the product of matrices B and A, denoted as $\Delta W = BA$, is zero initially [14].

The important hyperparameters such as the num_train_epochs, learning_rate, lora_rank, lora_alpha, and target_modules were thoroughly examined with multiple settings, which concluded with the conduct of 16 experiments to deduce the final model. The parameter set is as follows:

1) num_train_epochs = 100
2) learning_rate = 1e – 4
3) lora_rank = 16
4) lora_alpha = 32
5) target_modules = ["q_proj", "k_proj", "v_proj"]

Note that even though the training epochs were initially set to 100, the model training took place only for 43 epochs as the initiated early stopping mechanism terminated the procedure.

In this study, the author conducted fine-tuning experiments using three distinct pre-trained transformers: BART, T5, and PEGASUS. Nevertheless, the investigations revealed that BART exhibits superior performance compared to the other chosen models.

### C. Data Preprocessing

Preprocessing the data before training the model was essential. The process from the initial raw data to various data transformations to the final implementation proceeded as follows.

1) The data was retrieved from the Hugging Face Datasets repository. At first, only the textile domain of the BigPatent dataset [2] was selected and stored in the DatasetDict format.
2) The data retrieved was split into three sets: 70% for training, 20% for testing, and 10% for validation. This partitioning was done using the Scikit-learn library.
3) The data was subsequently analyzed by examining the word and sentence counts of both the descriptions and their corresponding abstracts to determine the characteristics of the data.
4) Text preprocessing approaches that were undertaken are as follows:
   a) Lowercasing - Lowercase word conversion (NLP → nlp). When not converted to lowercase, terms like "Research" and "research" are treated as two separate words in the vector space model, even though they have the same meaning.
   b) Stopword Removal - Words like "the", "is", and "a" are usually left out because they are common and have little meaning.
   c) Tokenization - To enable further processing or analysis, it entails dividing a text into smaller units or tokens, such as words or subwords.
   d) Padding and Truncation - These are methods used in natural language processing to ensure that the length of input sequences is consistent.

[1] https://huggingface.co/Nevidu/LexBartLo_1
[2] https://huggingface.co/Nevidu/LexBartLo_2

For the few-shot learning experiment, data was acquired from three distinct patent domains: human necessities, fixed constructions, and mechanical engineering. A total of 10 samples were gathered from each domain and combined with the existing training data. Additionally, 500 samples were collected from each domain for testing purposes.

## IV. Testing & Evaluation

During the model-building process, the author conducted experiments with various pre-trained models and hyperparameters to assess the performance of the developed hybrid architecture. Four distinguishable metrics, namely ROUGE-N, ROUGE-L, BERTScore, and METEOR, were chosen to assess the performance of these text summarizer models whilst the training and validation loss curve was retrieved to analyze.

Two specific splits were utilized to evaluate these models, containing patent articles and corresponding summaries from the domains of Textiles, Human Necessities, Fixed Constructions, and Mechanical Engineering. Split 1 consists of 2259 samples from the Textile domain, while Split 2 comprises samples from all four selected domains, totalling 2259 samples again.

A total of seventeen experiments were conducted, each of which focused on textile patent documents and made use of the Split_1 data. The best model was selected out of them to generalize the model. Those two models are detailed below, along with a discussion of the results.

TABLE I. DESCRIPTION OF IMPLEMENTED MODELS

| Model ID | Description |
|---|---|
| LexBartLo_1[1] | LexRank was used to rank the sentences, while LoRA was used to fine-tune the abstractive text summarization model, which was a pre-trained BART LARGE model. |
| LexBartLo_2[2] | LexRank was used to rank the sentences, while LoRA was used to fine-tune the abstractive text summarization model, which was a pre-trained BART LARGE model. (10 samples each from the disciplines of mechanical engineering, fixed construction, and human necessities were used for training.) |

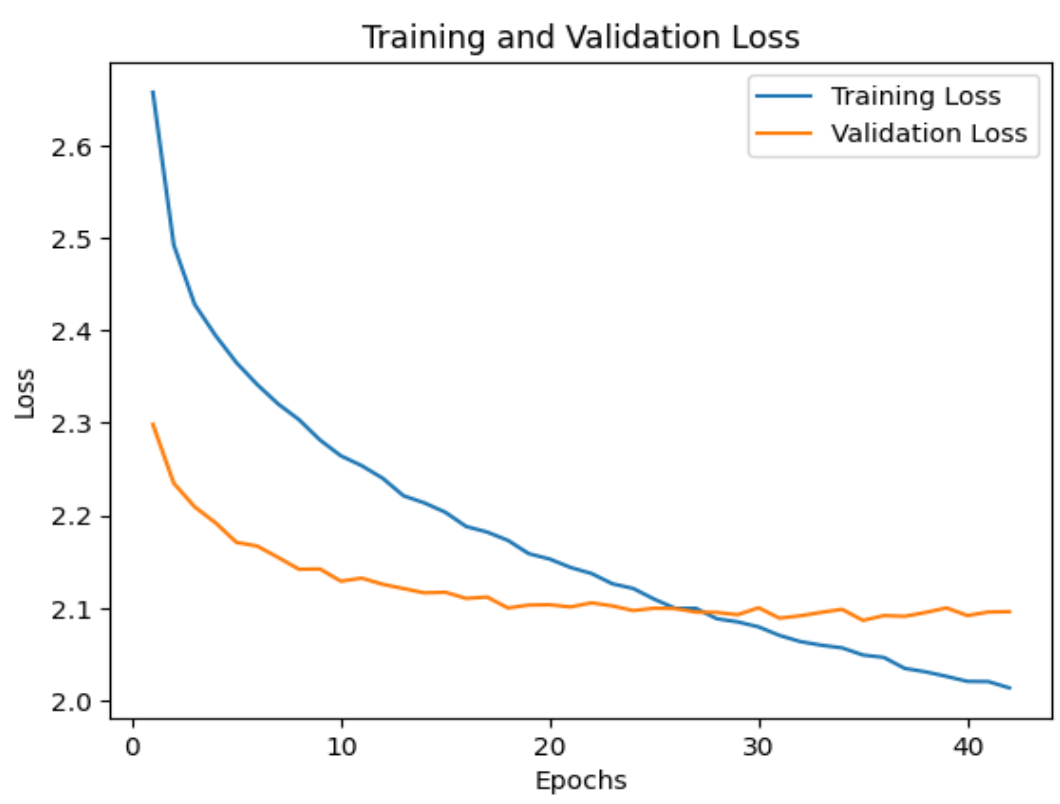

Fig. 3. Training and validation loss progression of model LexBartLo_1 across training epochs, illustrating model convergence.

The loss curve seen during training, which included the use of a separate validation set of 1130 samples, showed signs of overfitting. An early stopping mechanism was developed with a patience of seven, just monitoring the training loss, which continued to show signs of the problem. Nonetheless, it preemptively terminated the training process, emphasizing the importance of the technique used.

The author used benchmark models based on their performance as established by literature and available sources to compare the model suggested in this research to the same evaluation metrics used.

The table below provides information on the chosen models fine-tuned on the BigPatent dataset [2].

TABLE II. DESCRIPTION OF BENCHMARKING MODELS

| Model ID | Description |
|---|---|
| MT5 [15] | A multilingual variation of T5 was pre-trained using the exclusive dataset Common Crawl, which included 101 distinct languages. |
| PEGASUS [16] | Pegasus' pretraining task is purposefully built for summarization. It entails eliminating or masking essential sentences from an input text and generating them into an individual output sequence utilizing the remainder of phrases, equivalent to an extractive summary. |
| BigBirdPegasus [8] | BigBirdPegasus is a sparse-attention-based transformer created by the Google research team that can handle much longer sequences. BigBird uses not just sparse attention, but also global and random attention mechanisms in the input sequence. |
| LongT5 [7] | An encoder-decoder transformer that has been pre-trained in a generative text-to-text denoising context. The LongT5 model is an improved version of the T5 model that includes two unique and effective attention mechanisms: local attention and transient-global attention. |

The table below presents the evaluation results of the chosen benchmark models and the suggested models in this study.

TABLE III. BENCHMARKING RESULTS: ROUGE VARIANTS

| Model ID | ROUGE 1 | ROUGE 2 | ROUGE L | ROUGE LSum |
|---|---|---|---|---|
| LexBartLo_1 | 0.4576 | 0.2126 | 0.31472 | 0.31547 |
| LexBartLo_2 | 0.46325 | 0.21349 | 0.31338 | 0.31415 |
| MT5 | 0.1974 | 0.07916 | 0.16285 | 0.16307 |
| PEGASUS | 0.49066 | 0.2991 | 0.38775 | 0.38843 |
| BigBirdPegasus | **0.49959** | **0.3049** | **0.39188** | **0.39276** |
| LongT5 | 0.32772 | 0.10448 | 0.22204 | 0.22265 |

Recall-Oriented understudy for Gisting Evaluation (Rouge) is the most frequently employed similarity measure for text summarizing tasks; the technique includes several automatic evaluation methods, such as ROUGE-N, ROUGE -l, ROUGE -W, and ROUGE -S, which determine the degree of similarity in the summaries.

The most common of all the currently available ROUGE variants is the ROUGE-N, where n denotes the length of the n-gram, $gram_n$, and is the maximum number of n-grams co-occurring in both a candidate summary and a set of reference summaries [17]. In contrast, ROUGE-L utilizes the longest common subsequence (LCS) method to compare a generated summary with a reference summary.

The ROUGE-Lsum metric is closely linked to the ROUGE-L metric, although it has a slightly different calculating approach. The final score is computed by averaging all the results obtained using the ROUGE-L calculation procedure at the sentence level.

BERTSCORE evaluates summaries based on Cosine similarity but uses contextual embeddings of the words that promote its furtherance. Whereas the METEOR metric is calculated by taking the harmonic mean of the accuracy and recall for individual words, with a larger weight given to recall than precision [18].

TABLE IV. BENCHMARKING RESULTS: ADVANCED METRICS

| Model ID | BERTSCORE | METEOR |
|---|---|---|
| LexBartLo_1 | 0.87009 | 0.29591 |
| LexBartLo_2 | 0.86941 | 0.30468 |
| MT5 | 0.8431 | 0.08216 |
| PEGASUS | **0.88658** | 0.33027 |
| BigBirdPegasus | 0.88555 | **0.3607** |
| LongT5 | 0.85826 | 0.16517 |

A few-shot learning approach was proposed to provide knowledge of the other fields beyond textiles. Despite the inclusion of more samples during the training process for model LexBartLo_2, as previously stated, the performance difference is not substantial when compared to LexBartLo_1. This is to be anticipated given that the inclusion of just 30 samples does not significantly change the data distribution, which follows a convergence pattern like that of the LexBartLo_1 model trained only on Textiles domain data. Consequently, the author didn't perceive the inclusion of the remaining domains in the BigPatent dataset due to this similar convergence observed.

In terms of assessment measures, the BigBirdPegasus and Pegasus models slightly outperformed the author's proposed LexBartLo models, however the MT5 and LongT5 models were easily surpassed. Thus, the recommended model's abilities indicate that, despite the transformer models' not being fully fine-tuned, competitive performance is achievable.

## V. LIMITATIONS

The author identified a portion of limitations due to multiple constraints during the research, which necessitated certain sacrifices. The limits that demand attention are discussed in this segment.

Initially, the implemented model was exposed to only four specific technical classifications within the BigPatent dataset: Textiles, Human Necessities, Fixed Constructions, and Mechanical Engineering. However, the dataset also contains an additional five distinct technical domains that were not utilized in creating the generalized model. This occurred because it was not feasible to consistently test the model on a single test set that encompassed all nine domains within the given time frame. Consequently, the generalized model was not entirely applicable to all existing technical disciplines.

Although the research objectives were successfully achieved and the indicated research gap was addressed, the model was unable to surpass current state-of-the-art performances. The optimal hyperparameters could not be determined due to the requirement for significant hardware resources and time to carry out additional experiments.

Additionally, because of the significant hardware resources and extended training time needed, the author could not obtain progressively complex pre-trained transformer models that could function as the basis for the proposed hybrid architecture. Furthermore, the system does not prioritize user-provided keywords to facilitate the document summarizing process. Although integrating this functionality would have required several changes to the current system architecture, it would have greatly increased user engagement with the system and given the model a great deal of flexibility.

Resolving these identified shortcomings could improve the performance of the system and increase its relevance in real-world situations.

## VI. Future Work

The author encourages researchers with an interest in this area of study to take up the potential future enhancements suggested in this section.

The performance of the suggested model can be further improved by doing hyperparameter tuning, which determines the ideal hyperparameters to produce the optimum results. Furthermore, while evaluating the sentences in an input document, the suggested system's sentence ranking component disregards the contextual significance. Therefore, by offering contextual comprehension when calculating sentence similarity inside the procedures, it is possible to increase the effectiveness of this component. On the other hand, by using transformers with larger contextual window sizes, the model can be exposed to more text, which may improve its overall understanding of the input document.

In contrast, acquiring pre-trained transformers with much higher complexity could allow their superior contextual knowledge to be used to provide a more meaningful abstract summary. The system's flexibility can be further enhanced by using a reinforcement learning technique to get user feedback on improving the produced summary. To create a more relevant and customized summary to the user's preferences, the researchers may also focus on getting the targeted keywords that the user seeks to emphasize in the document.

Incorporating these enhancements could lead to noteworthy advancements in the domain of patent document summarization.

## VII. Conclusion

The author proposed a novel approach for generating abstractive text summaries of patent documents, significantly contributing to the field of study. The adopted approach faced multiple challenges, mainly due to the difficulty in achieving the optimal balance between resource consumption and performance trade-offs. Additionally, the author conducted thorough testing and assessment of the suggested system, providing valuable insights into the design's performance. Furthermore, this study identified and discussed limitations and potential future enhancements to encourage further research in this area domain.